\def\year{2017}\relax
\documentclass[letterpaper]{article}
\usepackage{aaai17}
\usepackage{times}
\usepackage{helvet}
\usepackage{courier}
\usepackage{amsfonts}
\usepackage{amsmath,amssymb,amscd,epsfig,amsfonts,rotating}
\usepackage{graphicx}
\usepackage{epstopdf}
\usepackage{subfigure}
\usepackage{multirow}
\usepackage{booktabs}
\usepackage{xcolor}
\usepackage{url}
\usepackage{amsmath,amscd,amsbsy,amssymb,latexsym,url,bm}
\usepackage{epsfig,epstopdf,graphicx,subfigure}
\usepackage{enumitem,balance}
\usepackage{wrapfig}
\usepackage{mathrsfs, euscript}
\usepackage{algorithm}
\usepackage{algorithmic}
\usepackage{amssymb}
\usepackage{ifpdf}
\usepackage{makecell}
\usepackage{tikz}
\usepackage{appendix}
\usetikzlibrary{matrix,chains,positioning,decorations.pathreplacing,arrows}

\newcommand{\bs}{\boldsymbol}
\newcommand{\softmax}{\text{softmax}}
\newcommand{\concat}{\oplus}

\usepackage{color}

\title{SeqGAN: Sequence Generative Adversarial Nets with Policy Gradient}
\author{
	Lantao Yu$^\dag$, Weinan Zhang$^\dag$\thanks{Weinan Zhang is the corresponding author.}, Jun Wang$^\ddag$, Yong Yu$^\dag$\\
	$^\dag$Shanghai Jiao Tong University, $^\ddag$University College London\\
	\{yulantao,wnzhang,yyu\}@apex.sjtu.edu.cn, j.wang@cs.ucl.ac.uk
}

\begin{document}

\maketitle
\begin{abstract}
As a new way of training generative models, Generative Adversarial Net (GAN) that uses a discriminative model to guide the training of the generative model has enjoyed considerable success in generating real-valued data. However, it has limitations when the goal is for generating sequences of discrete tokens. A major reason lies in that the discrete outputs from the generative model make it difficult to pass the gradient update from the discriminative model to the generative model. Also, the discriminative model can only assess a complete sequence, while for a partially generated sequence, it is non-trivial to balance its current score and the future one once the entire sequence has been generated. In this paper, we propose a sequence generation framework, called SeqGAN, to solve the problems. Modeling the data generator as a stochastic policy in reinforcement learning (RL), SeqGAN bypasses the generator differentiation problem by directly performing gradient policy update. The RL reward signal comes from the GAN discriminator judged on a complete sequence, and is passed back to the intermediate state-action steps using Monte Carlo search. Extensive experiments on synthetic data and real-world tasks demonstrate significant improvements over strong baselines.
\end{abstract}

\section{Introduction}
Generating sequential synthetic data that mimics the real one is an important problem in unsupervised learning. Recently, recurrent neural networks (RNNs) with long short-term memory (LSTM) cells \cite{hochreiter1997long} have shown excellent performance ranging from natural language generation to handwriting generation \cite{wen2015semantically,graves2013generating}. The most common approach to training an RNN is to maximize the log predictive likelihood of each true token in the training sequence given the previous observed tokens  \cite{salakhutdinov2009learning}.
However, as argued in \cite{bengio2015scheduled}, the maximum likelihood approaches suffer from so-called \emph{exposure bias} in the inference stage: the model generates a sequence iteratively and predicts next token conditioned on its previously predicted ones that may be never observed in the training data. Such a discrepancy between training and inference can incur accumulatively along with the sequence and will become prominent as the length of sequence increases.
To address this problem, \cite{bengio2015scheduled} proposed a training strategy called scheduled sampling (SS), where the generative model is partially fed with its own synthetic data as prefix (observed tokens) rather than the true data when deciding the next token in the training stage.
Nevertheless, \cite{huszar2015not} showed that SS is an inconsistent training strategy and fails to address the problem fundamentally. Another possible solution of the training/inference discrepancy problem is to build the loss function on the entire generated sequence instead of each transition. For instance, in the application of machine translation, a task specific sequence score/loss, bilingual evaluation understudy (BLEU) \cite{papineni2002bleu}, can be adopted to guide the sequence generation. However, in many other practical applications, such as poem generation \cite{zhang2014chinese} and chatbot \cite{hingston2009turing}, a task specific loss may not be directly available to score a generated sequence accurately.

General adversarial net (GAN) proposed by \cite{goodfellow2014generative} is a promising framework for alleviating the above problem.
Specifically, in GAN a discriminative net $D$ learns to distinguish whether a given data instance is real or not, and a generative net $G$ learns to confuse $D$ by generating high quality data.
This approach has been successful and been mostly applied in computer vision tasks of generating samples of natural images \cite{denton2015deep}.

Unfortunately, applying GAN to generating sequences has two problems. Firstly, GAN is designed for generating real-valued, continuous data but has difficulties in directly generating sequences of discrete tokens, such as texts \cite{huszar2015not}.
The reason is that in GANs, the generator starts with random sampling first and then a determistic transform, govermented by the model parameters. As such, the gradient of the loss from $D$ w.r.t. the outputs by $G$ is used to guide the generative model $G$ (paramters)  to slightly change the generated value to make it more realistic. If the generated data is based on discrete tokens, the ``slight change" guidance from the discriminative net makes little sense because there is probably no corresponding token for such slight change in the limited dictionary space \cite{goodfellow2016reddit}. Secondly, GAN can only give the score/loss for an entire sequence when it has been generated; for a partially generated sequence, it is non-trivial to balance how good as it is now and the future score as the entire sequence.



In this paper, to address the above two issues, we follow \cite{bachman2015data,bahdanau2016actor} and consider the sequence generation procedure as a sequential decision making process. The generative model is treated as an agent of reinforcement learning (RL); the state is the generated tokens so far and the action is the next token to be generated.
Unlike the work in \cite{bahdanau2016actor} that requires a task-specific sequence score, such as BLEU in machine translation, to give the reward, we employ a discriminator to evaluate the sequence and feedback the evaluation to guide the learning of the generative model. To solve the problem that the gradient cannot pass back to the generative model when the output is discrete, we regard the generative model as a stochastic parametrized policy. In our policy gradient, we employ Monte Carlo (MC) search to approximate the state-action value.
We directly train the policy (generative model) via policy gradient \cite{sutton1999policy}, which naturally avoids the differentiation difficulty for discrete data in a conventional GAN.


Extensive experiments based on synthetic and real data are conducted to investigate the efficacy and properties of the proposed SeqGAN. In our synthetic data environment, SeqGAN significantly outperforms the maximum likelihood methods, scheduled sampling and PG-BLEU. In three real-world tasks, i.e. poem generation, speech language generation and music generation, SeqGAN significantly outperforms the compared baselines in various metrics including human expert judgement.

\section{Related Work}


Deep generative models have recently drawn significant attention, and the ability of learning over large (unlabeled) data endows them with more potential and vitality \cite{salakhutdinov2009learning,bengio2013generalized}.
\cite{hinton2006fast} first proposed to use the contrastive divergence algorithm to efficiently training deep belief nets (DBN). \cite{bengio2013generalized} proposed denoising autoencoder (DAE) that learns the data distribution in a supervised learning fashion. Both DBN and DAE learn a low dimensional representation (encoding) for each data instance and generate it from a decoding network.
Recently, variational autoencoder (VAE) that combines deep learning with statistical inference intended to represent a data instance in a latent hidden space \cite{kingma2014auto}, while still utilizing (deep) neural networks for non-linear mapping. The inference is done via variational methods. All these generative models are trained by maximizing (the lower bound of) training data likelihood, which, as mentioned by \cite{goodfellow2014generative}, suffers from the difficulty of approximating intractable probabilistic computations.

\cite{goodfellow2014generative} proposed an alternative training methodology to generative models, i.e. GANs, where the training procedure is a \emph{minimax} game between a generative model and a discriminative model. This framework bypasses the difficulty of maximum likelihood learning and has gained striking successes in natural image generation \cite{denton2015deep}.
However, little progress has been made in applying GANs to sequence discrete data generation problems, e.g. natural language generation \cite{huszar2015not}. This is due to the generator network in GAN is designed to be able to adjust the output continuously, which does not work on discrete data generation \cite{goodfellow2016reddit}.


On the other hand, a lot of efforts have been made to generate structured sequences. Recurrent neural networks can be trained to produce sequences of tokens in many applications such as machine translation \cite{sutskever2014sequence,bahdanau2014neural}. The most popular way of training RNNs is to maximize the likelihood of each token in the training data whereas \cite{bengio2015scheduled} pointed out that the discrepancy between training and generating makes the maximum likelihood estimation suboptimal and proposed scheduled sampling strategy (SS).
Later \cite{huszar2015not} theorized that the objective function underneath SS is improper and explained the reason why GANs tend to generate natural-looking samples in theory. Consequently, the GANs have great potential but are not practically feasible to discrete probabilistic models currently.

As pointed out by \cite{bachman2015data}, the sequence data generation can be formulated as a sequential decision making process, which can be potentially be solved by reinforcement learning techniques. Modeling the sequence generator as a policy of picking the next token, policy gradient methods \cite{sutton1999policy} can be adopted to optimize the generator once there is an (implicit) reward function to guide the policy.
For most practical sequence generation tasks, e.g. machine translation \cite{sutskever2014sequence}, the reward signal is meaningful only for the entire sequence, for instance in the game of Go  \cite{silver2016mastering}, the reward signal is only set at the end of the game. In those cases, state-action evaluation methods such as Monte Carlo (tree) search have been adopted \cite{browne2012survey}. By contract, our proposed SeqGAN extends GANs with the RL-based generator to solve the sequence generation problem, where a reward signal is provided by the discriminator at the end of each episode via Monte Carlo approach, and the generator picks the action and learns the policy using estimated overall rewards.



\section{Sequence Generative Adversarial Nets}
The sequence generation problem is denoted as follows. Given a dataset of real-world structured sequences, train a $\theta$-parameterized generative model $G_\theta$ to produce a sequence $Y_{1:T} = (y_1,\ldots,y_t,\ldots,y_T),y_t \in \mathcal{Y}$, where $\mathcal{Y}$ is the vocabulary of candidate tokens. We interpret this problem based on reinforcement learning. In timestep $t$, the state $s$ is the current produced tokens $(y_1,\ldots,y_{t-1})$ and the action $a$ is the next token $y_t$ to select. Thus the policy model $G_\theta(y_t|Y_{1:t-1})$  is stochastic, whereas the state transition is deterministic after an action has been chosen, i.e. $\delta_{s,s'}^a=1$ for the next state $s'= Y_{1:t}$ if the current state $s = Y_{1:t-1}$ and the action $a= y_t$; for other next states $s''$, $\delta_{s,s''}^a=0$.

\begin{figure}[t]

\vspace{-4pt}
\centering
\includegraphics[width=0.45\textwidth]{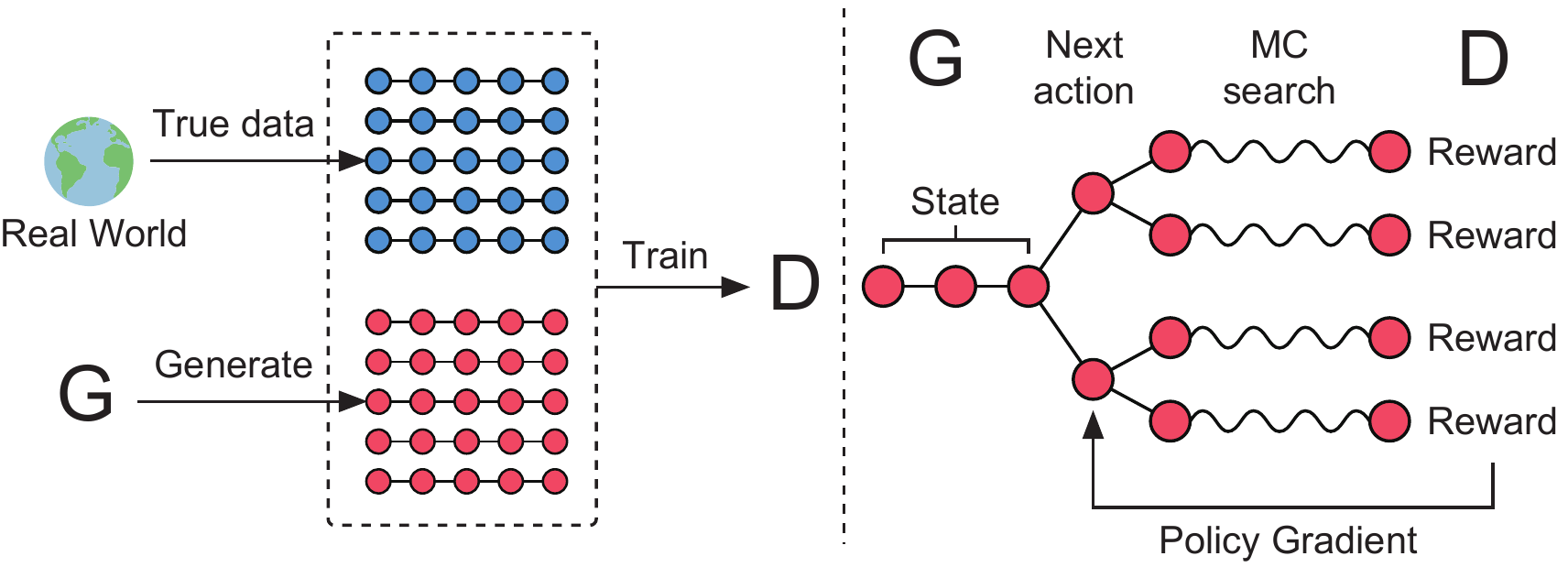}
\vspace{-8pt}
\caption{The illustration of SeqGAN.
	Left: $D$ is trained over the real data and the generated data by $G$.
	Right: $G$ is trained by policy gradient where the final reward signal is provided by $D$ and is passed back to the intermediate action value via Monte Carlo search.}
\label{fig:illustration_of_genGan}
\vspace{-10pt}
\end{figure}

Additionally, we also train a $\phi$-parameterized discriminative model $D_\phi$ \cite{goodfellow2014generative} to provide a guidance for improving generator $G_\theta$. $D_\phi(Y_{1:T})$ is a probability indicating how likely a sequence $Y_{1:T}$ is from real sequence data or not. As illustrated in Figure \ref{fig:illustration_of_genGan}, the discriminative model $D_\phi$ is trained by providing positive examples from the real sequence data and negative examples from the synthetic sequences generated from the generative model $G_\theta$. At the same time,  the generative model $G_\theta$ is updated by employing a policy gradient and MC search on the basis of the expected end reward received from the discriminative model $D_\phi$. The reward is estimated by the likelihood that it would fool the discriminative model $D_\phi$. The specific formulation is given in the next subsection.



\subsection{SeqGAN via Policy Gradient}
Following \cite{sutton1999policy}, when there is no intermediate reward, the objective of the generator model (policy) $G_\theta(y_t|Y_{1:t-1})$
is to generate a sequence from the start state $s_0$ to maximize its expected end reward:
{\small\begin{align}
J(\theta) & = \mathbb{E}[R_T|s_0,\theta] =\sum_{y_1 \in \mathcal{Y}} G_\theta(y_1|s_0) \cdot Q^{G_\theta}_{D_\phi}(s_0, y_1), \label{eq:g-obj}
\end{align}}
where $R_T$ is the reward for a complete sequence. Note that the reward is from the discriminator $D_\phi$, which we will discuss later. $Q^{G_\theta}_{D_\phi}(s,a)$ is the action-value function of a sequence, i.e. the expected accumulative reward starting from state $s$, taking action $a$, and then following policy $G_\theta$. The rational of the objective function for a sequence is that starting from a given initial state, the goal of the generator is to generate a sequence which would make the discriminator consider it is real.


The next question is how to estimate the action-value function.
In this paper, we use the REINFORCE algorithm \cite{williams1992simple} and consider the estimated probability of being real by the discriminator $D_\phi(Y^n_{1:T})$ as the reward. Formally, we have:
{\small\begin{equation}
Q^{G_\theta}_{D_\phi}(a=y_T,s=Y_{1:T-1}) = D_\phi(Y_{1:T}).
\end{equation}}
However, the discriminator only provides a reward value for a finished sequence. Since we actually care about the long-term reward, at every timestep, we should not only consider the fitness of previous tokens (prefix) but also the resulted future outcome. This is similar to playing the games such as Go or Chess where players sometimes would give up the immediate interests for the long-term victory \cite{silver2016mastering}.
Thus, to evaluate the action-value for an intermediate state, we apply Monte Carlo search with a roll-out policy $G_\beta$ to sample the unknown last $T-t$ tokens. We represent an $N$-time Monte Carlo search as
{\small\begin{equation}
\left\{Y^1_{1:T},\ldots,Y^N_{1:T}\right\} = \text{MC}^{G_\beta}(Y_{1:t};N),
\end{equation}}
where $Y^n_{1:t} = (y_1,\ldots,y_{t})$ and $Y^n_{t+1:T}$ is sampled based on the roll-out policy $G_\beta$ and the current state. In our experiment, $G_\beta$ is set the same as the generator, but one can use a simplified version if the speed is the priority \cite{silver2016mastering}.  To reduce the variance and get more accurate assessment of the action value, we run the roll-out policy starting from current state till the end of the sequence for $N$ times to get a batch of output samples. Thus, we have:
{\small
\begin{align}
& Q^{G_\theta}_{D_\phi}(s=Y_{1:t-1}, a=y_t) = \label{eq:q-mcs} \\
&\left\{
\begin{array}{lcl}
	\frac{1}{N} \sum_{n=1}^{N} D_\phi(Y^n_{1:T}),~Y^n_{1:T} \in \text{MC}^{G_\beta}(Y_{1:t};N) &\text{for}&{t<T}\\
	 D_\phi(Y_{1:t}) &\text{for}&{t=T},
	\end{array}
\right. \nonumber
\end{align}}
where, we see that when no intermediate reward, the function is iteratively defined as the next-state value starting from state $s'=Y_{1:t}$ and  rolling out to the end.


A benefit of using the discriminator $D_\phi$ as a reward function is that it can be dynamically updated to further improve the generative model iteratively. Once we have a set of more realistic generated sequences, we shall re-train the discriminator model as follows:
{\small
\begin{equation}
\min_\phi -\mathbb{E}_{Y \sim p_{\text{data}}} [\log D_\phi(Y)] -\mathbb{E}_{Y \sim G_\theta} [\log(1-D_\phi(Y))].  \label{eq:d-obj}
\end{equation}}

Each time when a new discriminator model has been obtained, we are ready to update the generator. The proposed policy based method relies upon optimizing a parametrized policy to directly maximize the long-term reward. Following \cite{sutton1999policy}, the gradient of the objective function
$J(\theta)$ w.r.t. the generator's parameters $\theta$ can be derived as

\begin{align}
\resizebox{\columnwidth}{!}{
$\nabla_\theta J(\theta) = \sum_{t=1}^{T}\mathbb{E}_{Y_{1:t-1} \sim G_\theta} \big[\sum \limits_{y_t \in \mathcal{Y}} \nabla_\theta G_\theta(y_t|Y_{1:t-1}) \cdot Q^{G_\theta}_{D_\phi}(Y_{1:t-1},y_t) \big].$
}\label{eq:j-derivative}
\end{align}

The above form is due to the deterministic state transition and zero intermediate rewards. The detailed derivation is provided in the appendix.
Using likelihood ratios \cite{glynn1990likelihood,sutton1999policy}, we build an unbiased estimation for Eq.~(\ref{eq:j-derivative}) (on one episode):

\resizebox{.95\columnwidth}{!}{
\begin{minipage}{\columnwidth}
{\small
\begin{align}
& \nabla_\theta J(\theta) \simeq \sum_{t=1}^{T} \sum_{y_t \in \mathcal{Y}} \nabla_\theta G_\theta(y_t|Y_{1:t-1}) \cdot Q_{D_\phi}^{G_\theta}(Y_{1:t-1},y_t)\hspace{-5pt}\\
&= \sum_{t=1}^{T} \sum_{y_t \in \mathcal{Y}} G_\theta(y_t|Y_{1:t-1}) \nabla_\theta \log G_\theta(y_t|Y_{1:t-1}) \cdot Q_{D_\phi}^{G_\theta}(Y_{1:t-1},y_t) \nonumber \\
&= \sum_{t=1}^{T} \mathbb{E}_{y_t \sim G_\theta(y_t|Y_{1:t-1})} [\nabla_\theta \log G_\theta(y_t|Y_{1:t-1}) \cdot Q_{D_\phi}^{G_\theta}(Y_{1:t-1},y_t)], \nonumber
\end{align}}
\end{minipage}}

where $Y_{1:t-1}$ is the observed intermediate state sampled from $G_\theta$. Since the expectation $\mathbb{E}[\cdot]$ can be approximated by sampling methods, we then update the generator's parameters as:
{\small\begin{equation}
\theta \leftarrow \theta + \alpha_h \nabla_\theta J(\theta), \label{eq:pg}
\end{equation}}
where $\alpha_h \in \mathbb{R}^+$ denotes the corresponding learning rate at $h$-th step. Also the advanced gradient algorithms such as Adam and RMSprop  can be adopted here.

In summary, Algorithm~\ref{alg:framework} shows full details of the proposed SeqGAN. At the beginning of the training, we use the maximum likelihood estimation (MLE) to pre-train $G_\theta$ on  training set $\mathcal{S}$.
We found the supervised signal from the pre-trained discriminator is informative to help adjust the generator efficiently.


After the pre-training, the generator and discriminator are trained alternatively. As the generator gets progressed via training on g-steps updates, the discriminator needs to be re-trained periodically to keeps a good pace with the generator.
When training the discriminator, positive examples are from the given dataset $\mathcal{S}$, whereas negative examples are generated from our generator. In order to keep the balance, the number of negative examples we generate for each d-step is the same as the positive examples. And to reduce the variability of the estimation, we use different sets of negative samples combined with positive ones, which is similar to bootstrapping \cite{quinlan1996bagging}.

\begin{algorithm}[t]
\caption{Sequence Generative Adversarial Nets}\label{alg:framework}
\begin{algorithmic}[1]
\small
\REQUIRE
generator policy $G_\theta$; roll-out policy $G_\beta$; discriminator $D_\phi$; a sequence dataset $\mathcal{S}=\left\{X_{1:T}\right\}$

\STATE
Initialize $G_\theta$, $D_\phi$ with random weights $\theta,\phi$.
\STATE
Pre-train $G_\theta$ using MLE on $\mathcal{S}$
\STATE
$\beta \gets \theta$
\STATE
Generate negative samples using $G_\theta$ for training $D_\phi$
\STATE
Pre-train $D_\phi$ via minimizing the cross entropy
\REPEAT
\FOR{g-steps}
\STATE
Generate a sequence $Y_{1:T}=(y_1,\ldots,y_T) \sim G_\theta$
\FOR{$t$ in $1:T$}
\STATE
Compute $Q(a=y_t; s=Y_{1:t-1})$ by Eq.~(\ref{eq:q-mcs})
\ENDFOR
\STATE
Update generator parameters via policy gradient Eq.~(\ref{eq:pg})
\ENDFOR
\FOR{d-steps}
\STATE
Use current $G_\theta$ to generate negative examples and combine with given positive examples $\mathcal{S}$
\STATE
Train discriminator $D_\phi$ for $k$ epochs by Eq.~({\ref{eq:d-obj}})
\ENDFOR
\STATE
$\beta \gets \theta$
\UNTIL{SeqGAN converges}
\end{algorithmic}
\end{algorithm}

\subsection{The Generative Model for Sequences}
We use recurrent neural networks (RNNs) \cite{hochreiter1997long} as the generative model. An RNN maps the input embedding representations $\bs{x}_1,\ldots,\bs{x}_T$ of the sequence $x_1,\ldots,x_T$ into a sequence of hidden states $\bs{h}_1,\ldots,\bs{h}_T$ by using the update function $g$ recursively.
{\small\begin{equation}
\bs{h}_t = g(\bs{h}_{t-1},\bs{x}_t)\label{eq:update-function}
\end{equation}}
Moreover, a softmax output layer $z$ maps the hidden states into the output token distribution
{\small\begin{equation}
p(y_t|x_1,\ldots,x_t) = z(\bs{h}_t) = \softmax(\bs{c}+\bs{Vh}_t),
\end{equation}}
where the parameters are a bias vector $\bs{c}$ and a weight matrix $\bs{V}$.
To deal with the common vanishing and exploding gradient problem \cite{goodfellow2016deep} of the backpropagation through time, we leverage the Long Short-Term Memory (LSTM) cells \cite{hochreiter1997long} to implement the update function $g$ in Eq.~(\ref{eq:update-function}).
It is worth noticing that most of the RNN variants, such as the gated recurrent unit (GRU) \cite{cho2014learning} and soft attention mechanism \cite{bahdanau2014neural}, can be used as a generator in SeqGAN.




\subsection{The Discriminative Model for Sequences}

Deep discriminative models such as deep neural network (DNN) \cite{vesely2013sequence}, convolutional neural network (CNN) \cite{kim2014convolutional} and recurrent convolutional neural network (RCNN) \cite{lai2015recurrent} have shown a high performance in complicated sequence classification tasks. In this paper, we choose the CNN as our discriminator as CNN has recently been shown of great effectiveness in text (token sequence) classification \cite{zhang2015text}.
Most discriminative models can only perform classification well for an entire sequence rather than the unfinished one. In this paper, we also focus on the situation where the discriminator predicts the probability that a finished sequence is real.\footnote{In our work, the generated sequence has a fixed length $T$, but note that CNN is also capable of the variable-length sequence discrimination with the max-over-time pooling technique \cite{kim2014convolutional}.}


We first represent an input sequence $x_1,\ldots,x_T$ as:
{\small\begin{equation}
\mathcal{E}_{1:T} = \bs{x}_1 \concat \bs{x}_2 \concat \ldots \concat \bs{x}_T,
\end{equation}}
where $\bs{x}_t\in \mathbb{R}^k$ is the $k$-dimensional token embedding and $\concat$ is the concatenation operator to build the matrix $\mathcal{E}_{1:T} \in \mathbb{R}^{T\times k}$. Then a kernel $\bs{w} \in \mathbb{R}^{l \times k}$ applies a convolutional operation to a window size of $\mathnormal{l}$ words to produce a new feature map:
{\small\begin{equation}
c_i = \rho(\bs{w}\otimes \mathcal{E}_{i:i+l-1}+b),
\end{equation}}
where $\otimes$ operator is the summation of elementwise production, $\mathnormal{b}$ is a bias term and $\rho$ is a non-linear function.
We can use various numbers of kernels with different window sizes to extract different features.
Finally we apply a max-over-time pooling operation over the feature maps $\tilde{c} = \max\left\{c_1,\ldots,c_{T-l+1}\right\}$.

To enhance the performance, we also add the highway architecture \cite{srivastava2015highway} based on the pooled feature maps. Finally, a fully connected layer with sigmoid activation is used to output the probability that the input sequence is real.
The optimization target is to minimize the cross entropy between the ground truth label and the predicted probability as formulated in Eq.~(\ref{eq:d-obj}).

Detailed implementations of the generative and discriminative models are provided in the appendix.

\section{Synthetic Data Experiments}
To test the efficacy and add our understanding  of SeqGAN, we conduct a simulated test with synthetic data\footnote{
Experiment code: {\scriptsize \url{https://github.com/LantaoYu/SeqGAN}}}. To simulate the real-world structured sequences, we consider a language model to capture the dependency of the tokens. We use a randomly initialized LSTM as the true model, aka, the oracle, to generate the real data distribution $p(x_t|x_1,\ldots,x_{t-1})$ for the following experiments.

\subsection{Evaluation Metric}
The benefit of having such oracle is that firstly, it provides the training dataset and secondly evaluates the exact performance of the generative models, which will not be possible with real data.  We know that MLE is trying to minimize the cross-entropy between the true data distribution $p$ and our approximation $q$, i.e. $-\mathbb{E}_{x \sim p}\log q(x)$.
However, the most accurate way of evaluating generative models is that we draw some samples from it and let human observers review them based on their prior knowledge. We assume that the human observer has learned an accurate model of the natural distribution $p_{\text{human}}(x)$. Then in order to increase the chance of passing Turing Test, we actually need to minimize the exact opposite average negative log-likelihood $-\mathbb{E}_{x \sim q}\log p_{\text{human}}(x)$ \cite{huszar2015not}, with the role of $p$ and $q$ exchanged. In our synthetic data experiments, we can consider the oracle to be the human observer for real-world problems, thus a perfect evaluation metric should be
{\small\begin{align}
\text{NLL}_{\text{oracle}}=-\mathbb{E}_{Y_{1:T} \sim G_\theta} \Big[\sum_{t=1}^T\log G_{\text{oracle}}(y_t|Y_{1:t-1}) \Big],
\end{align}}
where $G_\theta$ and $G_{\text{oracle}}$ denote our generative model and the oracle respectively.

At the test stage, we use $G_\theta$ to generate 100,000 sequence samples and calculate $\text{NLL}_{\text{oracle}}$ for each sample by $G_{\text{oracle}}$ and their average score. Also significance tests are performed to compare the statistical properties of the generation performance between the baselines and SeqGAN.

\subsection{Training Setting}
To set up the synthetic data experiments, we first initialize the parameters of an LSTM network following the normal distribution $\mathcal{N}(0,1)$ as the oracle describing the real data distribution $G_{\text{oracle}}(x_t|x_1,\ldots,x_{t-1})$. Then we use it to generate 10,000 sequences of length 20 as the training set $\mathcal{S}$ for the generative models.

In SeqGAN algorithm, the training set for the discriminator is comprised by the generated examples with the label 0 and the instances from $\mathcal{S}$ with the label 1. For different tasks, one should design specific structure for the convolutional layer and in our synthetic data experiments, the kernel size is from 1 to $T$ and the number of each kernel size is between 100 to 200\footnote{Implementation details are in the appendix.}. Dropout \cite{srivastava2014dropout} and L2 regularization are used to avoid over-fitting.

Four generative models are compared with SeqGAN. The first model is a random token generation. The second one is the MLE trained LSTM $G_\theta$. The third one is scheduled sampling \cite{bengio2015scheduled}. The fourth one is the Policy Gradient with BLEU (PG-BLEU). In the scheduled sampling, the training process gradually changes from a fully guided scheme feeding the true previous tokens into LSTM, towards a less guided scheme which mostly feeds the LSTM with its generated tokens. A curriculum rate $\omega$ is used to control the probability of replacing the true tokens with the generated ones.
To get a good and stable performance, we decrease $\omega$ by 0.002 for every training epoch. In the PG-BLEU algorithm, we use BLEU, a metric measuring the similarity between a generated sequence and references (training data), to score the finished samples from Monte Carlo search.

\subsection{Results}

\begin{table}[t]
\centering
\caption{Sequence generation performance comparison. The $p$-value is between SeqGAN and the baseline from T-test.}\vspace{5pt}
\small
\label{tab:sequence-overall}
\resizebox{\columnwidth}{!}{
\begin{tabular}{c|c|c|c|c|c}
Algorithm & Random & MLE & SS & PG-BLEU & SeqGAN\\
\hline
NLL & 10.310 & 9.038 & 8.985 & 8.946 & \textbf{8.736} \\
\hline
$p$-value & $< 10^{-6}$ & $< 10^{-6}$ & $< 10^{-6}$ & $< 10^{-6}$ &
\end{tabular}
}
\vspace{-10pt}
\end{table}

\begin{figure}[t]
\vspace{10pt}
\centering
\includegraphics[width=0.8\columnwidth]{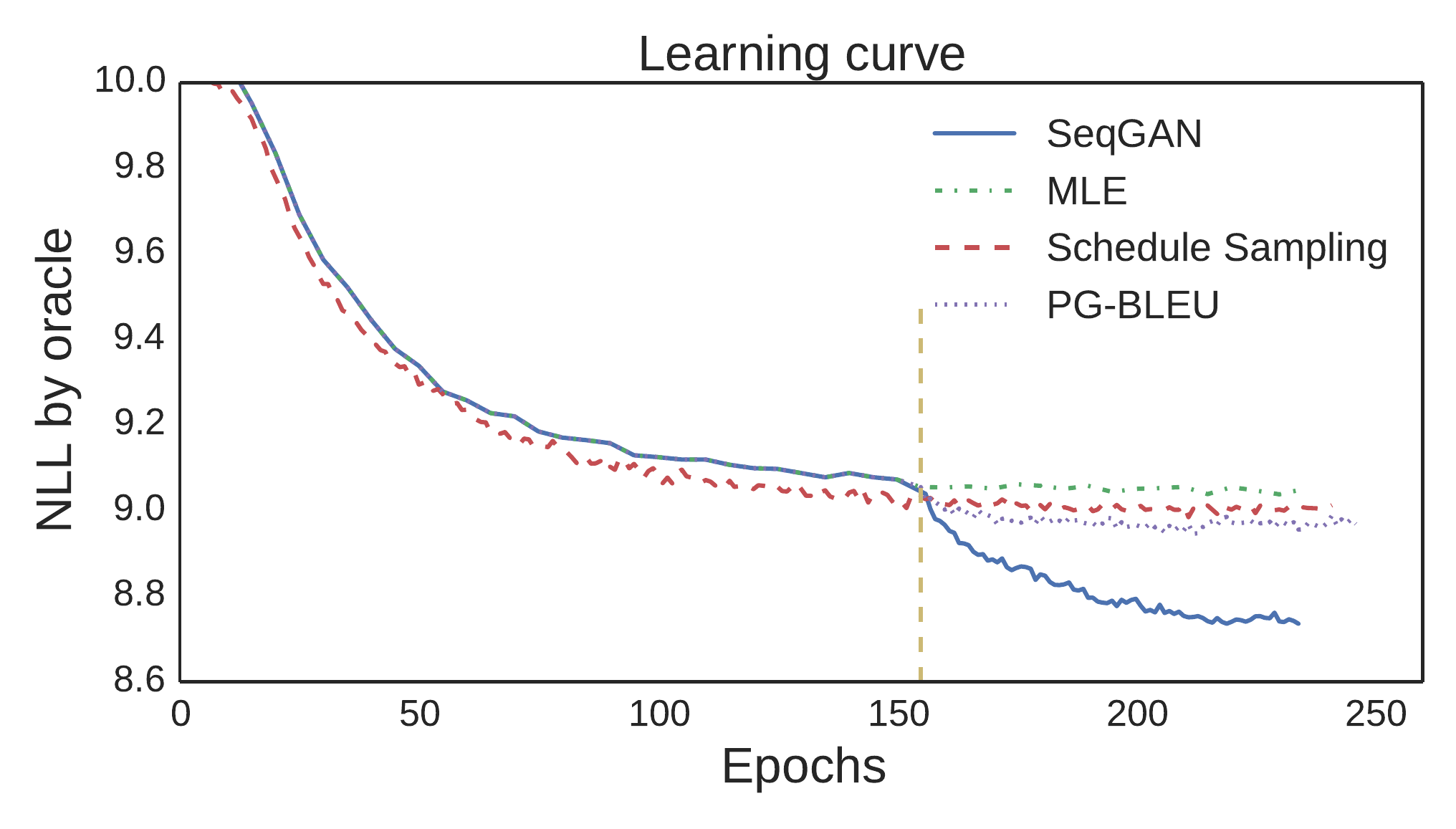}
\vspace{-12pt}
\caption{Negative log-likelihood convergence w.r.t. the training epochs. The vertical dashed line represents the end of pre-training for SeqGAN, SS and PG-BLEU.}
\label{fig:learning-curve}
\vspace{-10pt}
\end{figure}

The $\text{NLL}_{\text{oracle}}$ performance of generating sequences from the compared policies is provided in Table \ref{tab:sequence-overall}. Since the evaluation metric is fundamentally instructive, we can see the impact of SeqGAN, which outperforms other baselines significantly. A significance T-test on the $\text{NLL}_{\text{oracle}}$ score distribution of the generated sequences from the compared models is also performed, which demonstrates the significant improvement of SeqGAN over all compared models.

The learning curves shown in Figure \ref{fig:learning-curve} illustrate the superiority of SeqGAN explicitly. After about 150 training epochs, both the maximum likelihood estimation and the schedule sampling methods converge to a relatively high $\text{NLL}_{\text{oracle}}$ score, whereas SeqGAN can improve the limit of the generator with the same structure as the baselines significantly. This indicates the prospect of applying adversarial training strategies to discrete sequence generative models to breakthrough the limitations of MLE.
Additionally, SeqGAN outperforms PG-BLEU, which means the discriminative signal in GAN is more general and effective than a predefined score (e.g. BLEU) to guide the generative policy to capture the underlying distribution of the sequence data.

\subsection{Discussion}

\begin{figure}[t]
\centering
\vspace{-10pt}
\subfigure[{\scriptsize ~~g-steps=100, d-steps=1, $k$=10}]{
\includegraphics[height=1.2in]{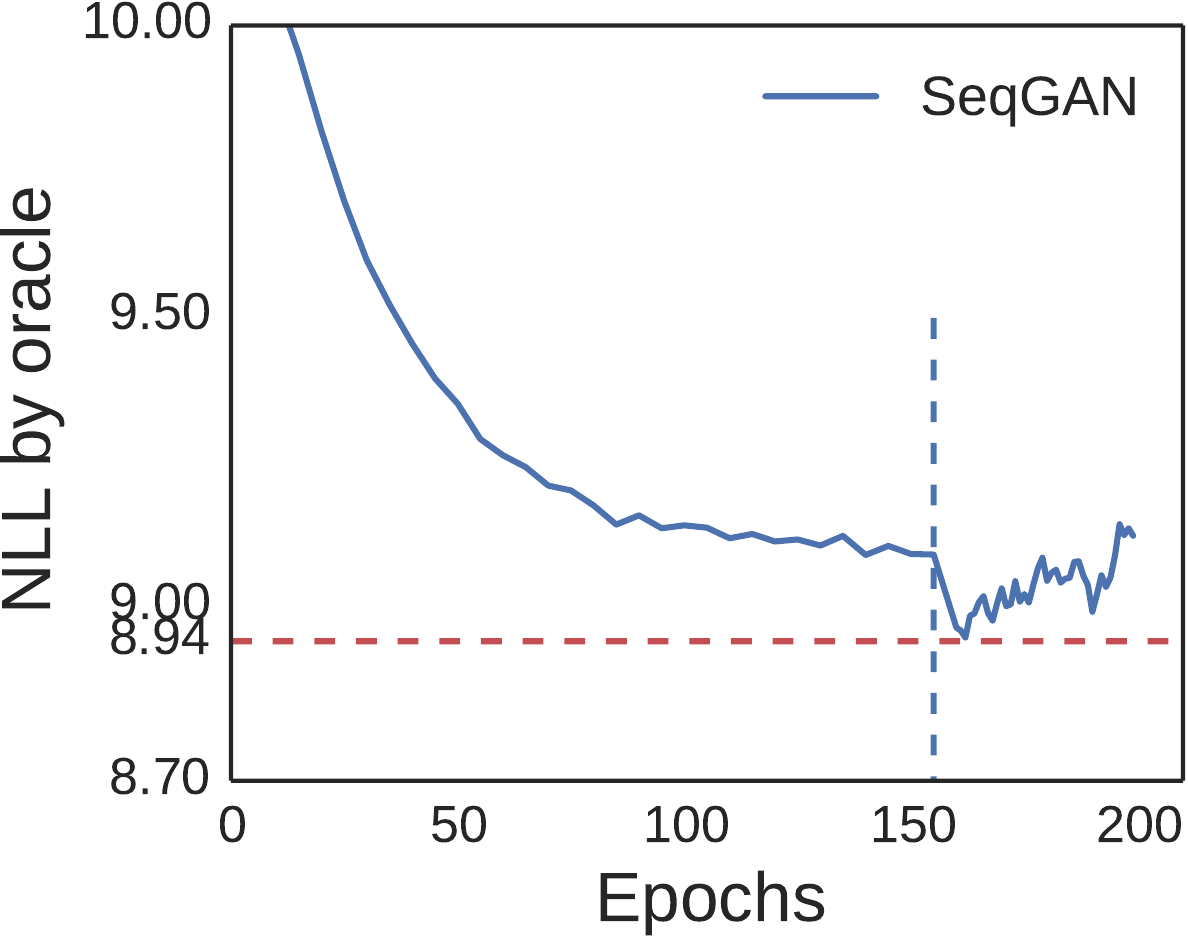}\label{fig:hyper-a}}
\subfigure[{\scriptsize ~~g-steps=30, d-steps=1, $k$=30}]{
\includegraphics[height=1.2in]{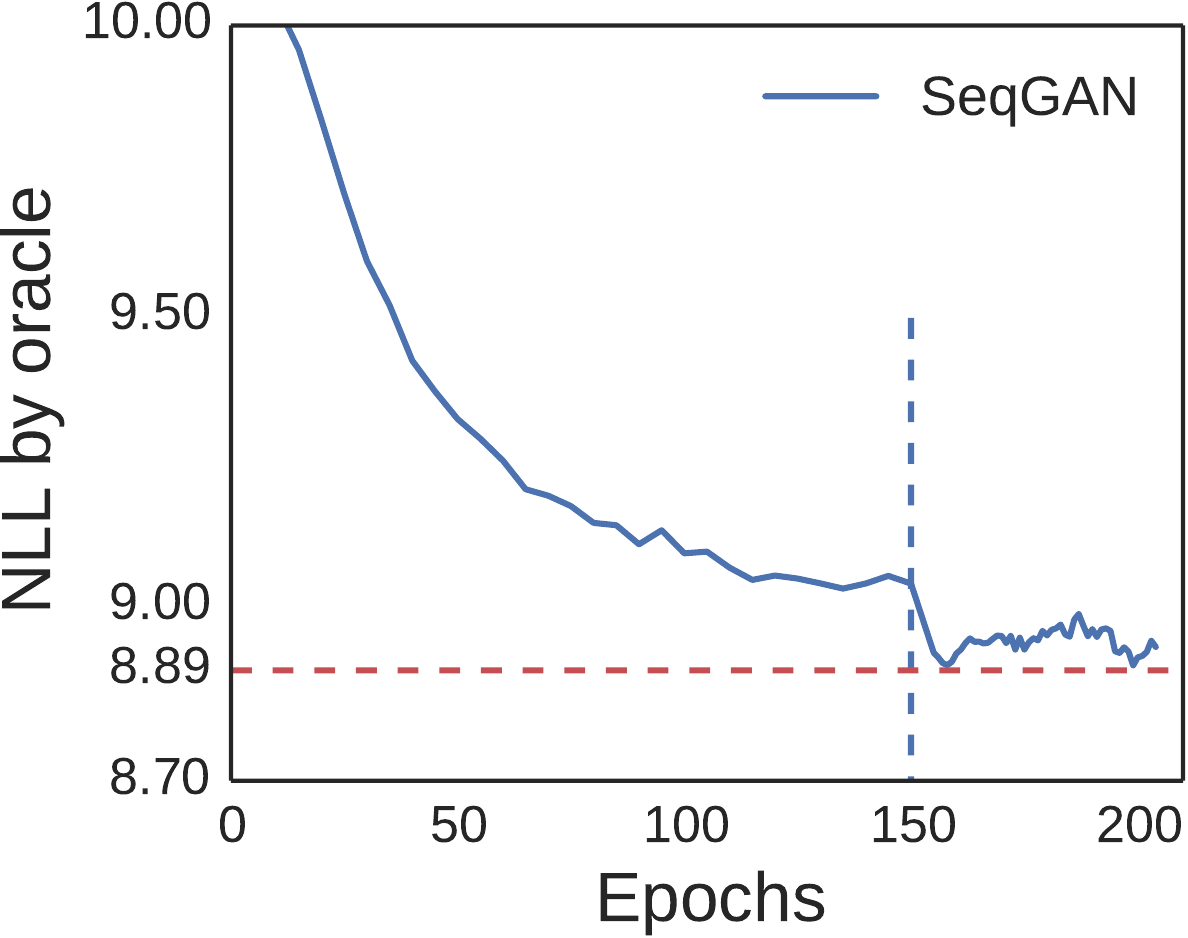}\label{fig:hyper-b}}
\subfigure[{\scriptsize ~~g-steps=1, d-steps=1, $k$=10}]{
\includegraphics[height=1.2in]{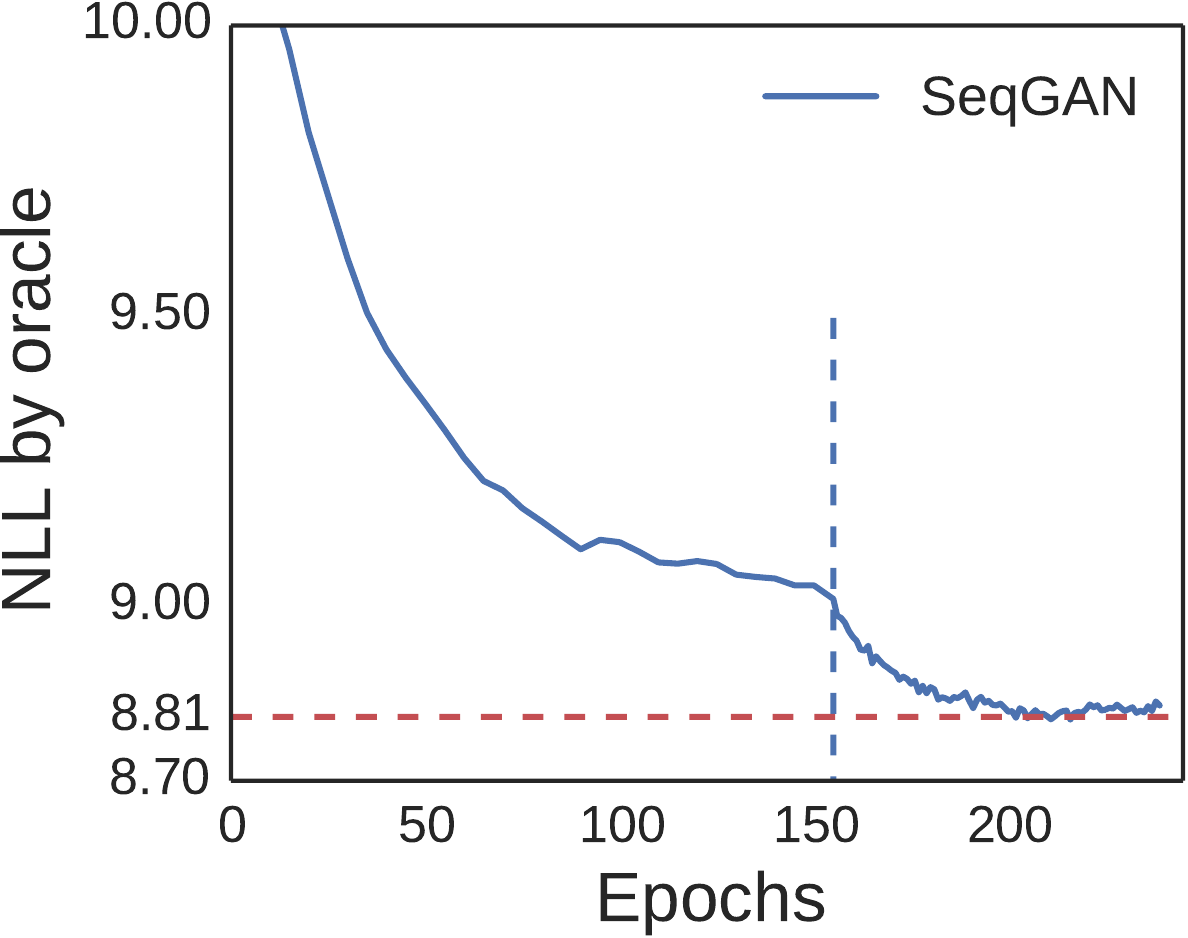}\label{fig:hyper-c}}
\subfigure[{\scriptsize ~~g-steps=1, d-steps=5, $k$=3}]{
\includegraphics[height=1.2in]{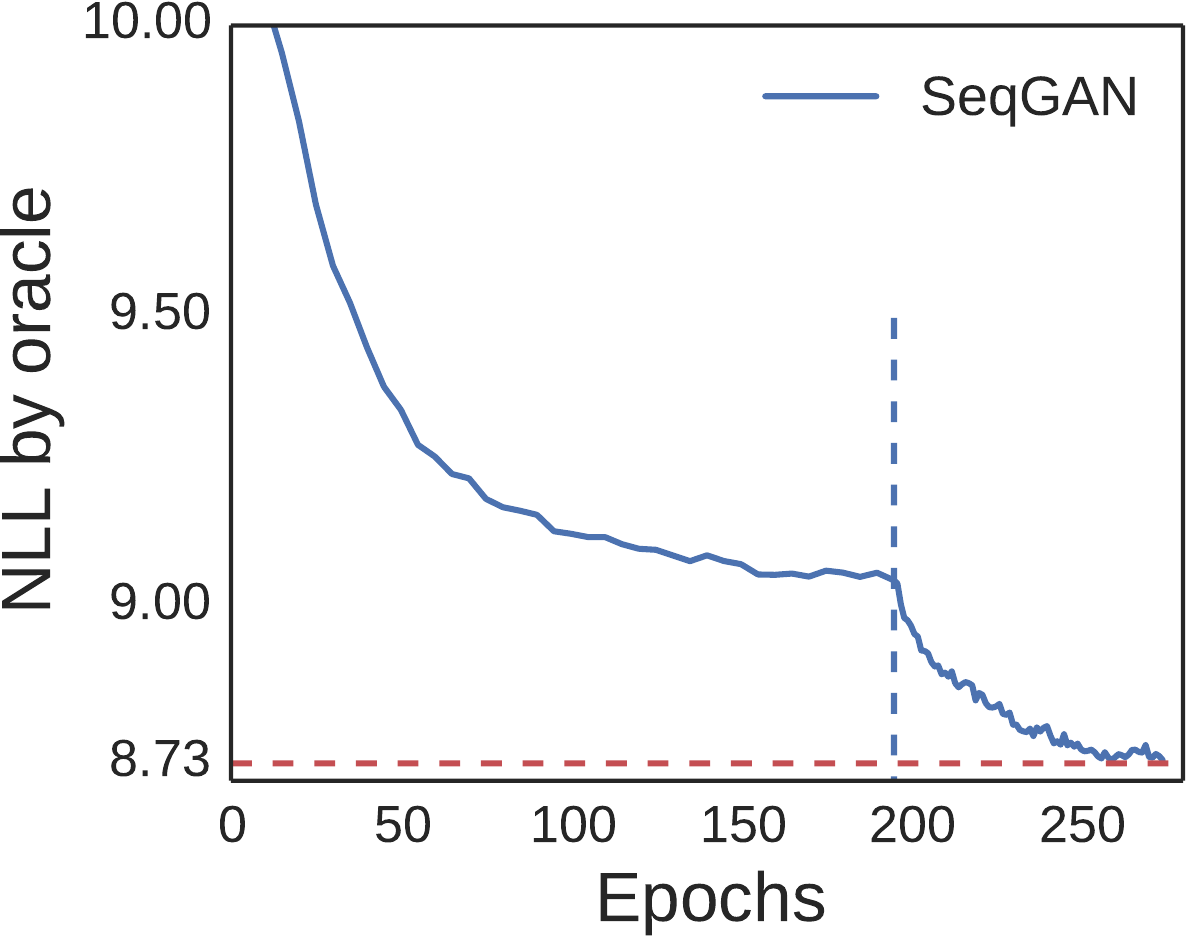}\label{fig:hyper-d}}
\vspace{-5pt}
\caption{Negative log-likelihood convergence performance of SeqGAN with different training strategies. The vertical dashed line represents the beginning of adversarial training.}
\label{fig:hyper}
\vspace{-10pt}
\end{figure}

In our synthetic data experiments, we find that the stability of SeqGAN depends on the training strategy. More specifically, the g-steps, d-steps and $k$ parameters in Algorithm~\ref{alg:framework} have a large effect on the convergence and performance of SeqGAN. Figure~\ref{fig:hyper} shows the effect of these parameters. In Figure~\ref{fig:hyper-a}, the g-steps is much larger than the d-steps and epoch number $k$, which means we train the generator for many times until we update the discriminator. This strategy leads to a fast convergence but as the generator improves quickly, the discriminator cannot get fully trained and thus will provide a misleading signal gradually. In Figure~\ref{fig:hyper-b}, with more discriminator training epochs, the unstable training process is alleviated. In Figure~\ref{fig:hyper-c}, we train the generator for only one epoch and then before the discriminator gets fooled, we update it immediately based on the more realistic negative examples. In such a case, SeqGAN learns stably.

The d-steps in all three training strategies described above is set to 1, which means we only generate one set of negative examples with the same number as the given dataset, and then train the discriminator on it for various $k$ epochs. But actually we can utilize the potentially unlimited number of negative examples to improve the discriminator. This trick can be considered as a type of bootstrapping,
where we combine the fixed positive examples with different negative examples to obtain multiple training sets. Figure \ref{fig:hyper-d} shows this technique can improve the overall performance with good stability, since the discriminator is shown more negative examples and each time the positive examples are emphasized, which will lead to a more comprehensive guidance for training generator. This is in line with the theorem in \cite{goodfellow2014generative}. When analyzing the convergence of generative adversarial nets, an important assumption is that the discriminator is allowed to reach its optimum given $G$. Only if the discriminator is capable of differentiating real data from unnatural data consistently, the supervised signal from it can be meaningful and the whole adversarial training process can be stable and effective.



\section{Real-world Scenarios}

To complement the previous experiments, we also test SeqGAN on several real-world tasks, i.e. poem composition, speech language generation and music generation.

\subsection{Text Generation}
For text generation scenarios, we apply the proposed SeqGAN to generate Chinese poems and Barack Obama political speeches.
In the poem composition task, we use a corpus\footnote{\scriptsize \url{http://homepages.inf.ed.ac.uk/mlap/Data/EMNLP14/}} of 16,394 Chinese quatrains, each containing four lines of twenty characters in total. To focus on a fully automatic solution and stay general, we did not use any prior knowledge of special structure rules in Chinese poems such as specific phonological rules. In the Obama political speech generation task, we use a corpus\footnote{\scriptsize \url{https://github.com/samim23/obama-rnn}}, which is a collection of 11,092 paragraphs from Obama's political speeches.

We use BLEU score as an evaluation metric to measure the similarity degree between the generated texts and the human-created texts.
BLEU is originally designed to automatically judge the machine translation quality \cite{papineni2002bleu}. The key point is to compare the similarity between the results created by machine and the references provided by human.
Specifically, for poem evaluation, we set n-gram to be 2 (BLEU-2) since most words (dependency) in classical Chinese poems consist of one or two characters \cite{yi2016generating} and for the similar reason, we use BLEU-3 and BLEU-4 to evaluate Obama speech generation performance. In our work, we use the whole test set as the references instead of trying to find some references for the following line given the previous line \cite{he2012generating}.
The reason is in generation tasks we only provide some positive examples and then let the model catch the patterns of them and generate new ones.
In addition to BLEU, we also choose poem generation as a case for human judgement since a poem is a creative text construction and human evaluation is ideal. Specifically, we mix the 20 real poems and 20 each generated from  SeqGAN and MLE. Then 70 experts on Chinese poems are invited to judge whether each of the 60 poem is created by human or machines. Once regarded to be real, it gets +1 score, otherwise 0. Finally, the average score for each algorithm is calculated.

The experiment results are shown in Tables \ref{tab:poem} and \ref{tab:speech}, from which we can see the significant advantage of SeqGAN over the MLE in text generation. Particularly, for poem composition, SeqGAN performs comparably to real human data.


\begin{table}[t]
	\centering
	\vspace{-10pt}
	\caption{Chinese poem generation performance comparison.}\label{tab:poem}
	\vspace{2pt}
	\small
	\begin{tabular}{c|c|c|c|c}
		Algorithm & Human score& $p$-value & BLEU-2 & $p$-value\\
		\hline
		MLE & 0.4165 & \multirow{2}*{0.0034} & 0.6670 & \multirow{2}*{$< 10^{-6}$} \\
		SeqGAN & \textbf{0.5356} & & \textbf{0.7389} &\\
		\hline
		Real data & 0.6011 & & 0.746 &
	\end{tabular}
	\vspace{-5pt}
	\caption{Obama political speech generation performance.}\label{tab:speech}
	\vspace{2pt}
	\small
	\begin{tabular}{c|c|c|c|c}
		Algorithm & BLEU-3& $p$-value & BLEU-4 & $p$-value\\
		\hline
		MLE & 0.519 & \multirow{2}*{$< 10^{-6}$} & 0.416 & \multirow{2}*{0.00014}\\
		SeqGAN & \textbf{0.556} &  & \textbf{0.427} &
	\end{tabular}

	\vspace{-5pt}
	\caption{Music generation performance comparison.}\label{tab:music}
	\vspace{2pt}
	\small
	\begin{tabular}{c|c|c|c|c}
		Algorithm & BLEU-4 & $p$-value & MSE & $p$-value\\
		\hline
		MLE & 0.9210 & \multirow{2}*{$< 10^{-6}$} & 22.38 &\multirow{2}*{0.00034}\\
		SeqGAN & \textbf{0.9406} & & \textbf{20.62}
	\end{tabular}
\vspace{-10pt}
\end{table}

\subsection{Music Generation}
For music composition, we use Nottingham\footnote{\scriptsize \url{http://www.iro.umontreal.ca/~lisa/deep/data}} dataset as our training data, which is a collection of 695 music of folk tunes in midi file format. We study the solo track of each music.
In our work, we use 88 numbers to represent 88 pitches, which correspond to the 88 keys on the piano. With the pitch sampling for every 0.4s\footnote{\scriptsize \url{http://deeplearning.net/tutorial/rnnrbm.html}}, we transform the midi files into sequences of numbers from 1 to 88 with the length 32.

To model the fitness of the discrete piano key patterns, BLEU is used as the evaluation metric. To model the fitness of the continuous pitch data patterns, the mean squared error (MSE) \cite{manaris2007corpus} is used for evaluation.

From Table \ref{tab:music}, we see that SeqGAN outperforms the MLE significantly in both metrics in the music generation task.

\section{Conclusion}
In this paper, we proposed a sequence generation method, SeqGAN, to effectively train generative adversarial nets for structured sequences generation via policy gradient. To our best knowledge, this is the first work extending GANs to generate sequences of discrete tokens. In our synthetic data experiments, we used an oracle evaluation mechanism to explicitly illustrate the superiority of SeqGAN over strong baselines. For three real-world scenarios, i.e., poems, speech language and music generation, SeqGAN showed excellent performance on generating the creative sequences. We also performed a set of experiments to investigate the robustness and stability of training SeqGAN. For future work, we plan to build Monte Carlo tree search and value network \cite{silver2016mastering} to improve action decision making for large scale data and in the case of longer-term planning.

\section{Acknowledgments}
We sincerely thank Tianxing He for many helpful discussions and comments on the manuscript.
{\small
\bibliography{references}
\bibliographystyle{aaai}}

\newpage
\appendix
\onecolumn

\begin{appendices}

\section{Appendix}
In Section 1, we present the step-by-step derivation of Eq.~(6) in the paper. In Section 2, the detailed realization of the generative model and the discriminative model is discussed, including the model parameter settings. In Section 3, an interesting ablation study is provided, which is a supplementary to the discussions of the synthetic data experiments.
~\\
~\\
~\\

\subsection{Proof for Eq.~(6)}
For readability, we provide the detailed derivation of Eq.~(6) here by following \cite{sutton1999policy}.

As mentioned in \textsc{Sequence Generative Adversarial Nets} section, the state transition is deterministic after an action has been chosen, i.e. $\delta_{s,s'}^a=1$ for the next state $s'= Y_{1:t}$ if the current state $s = Y_{1:t-1}$ and the action $a= y_t$; for other next states $s''$, $\delta_{s,s''}^a=0$. In addition, the intermediate reward $\mathcal{R}_s^a$ is 0. We re-write the action value and state value as follows:
\begin{align}
	& Q^{G_\theta}(s=Y_{1:t-1},a=y_t)=\mathcal{R}_s^a + \sum_{s' \in S} \delta^a_{ss'}V^{G_\theta}(s')=V^{G_\theta}(Y_{1:t})\\
	& V^{G_\theta}(s=Y_{1:t-1}) = \sum_{y_{t} \in \mathcal{Y}} G_\theta(y_t|Y_{1:t-1}) \cdot Q^{G_\theta}(Y_{1:t-1}, y_t)
\end{align}

For the start state $s_0$, the value is calculated as
\begin{align}
V^{G_\theta}(s_0) &= \mathbb{E}[R_T|s_0,\theta]\\
& = \sum_{y_1 \in \mathcal{Y}} G_\theta(y_1|s_0) \cdot Q^{G_\theta}(s_0,y_1),\nonumber
\end{align}
which is the objective function $J(\theta)$ to maximize in Eq.~(1) of the paper.

Then we can obtain the gradient of the objective function, defined in Eq.~(1), w.r.t. the generator's parameters $\theta$:
{\small
\begin{align}
& \nabla_\theta J(\theta) \nonumber\\
&= \nabla_\theta V^{G_\theta}(s_0) = \nabla_\theta [\sum_{y_1 \in \mathcal{Y}} G_\theta(y_1|s_0) \cdot Q^{G_\theta}(s_0,y_1)] \nonumber\\
& = \sum_{y_1 \in \mathcal{Y}} [\nabla_\theta G_\theta(y_1|s_0) \cdot Q^{G_\theta}(s_0,y_1)+G_\theta(y_1|s_0) \cdot \nabla_\theta Q^{G_\theta}(s_0,y_1)] \nonumber\\
& = \sum_{y_1 \in \mathcal{Y}} [\nabla_\theta G_\theta(y_1|s_0) \cdot Q^{G_\theta}(s_0,y_1)+G_\theta(y_1|s_0) \cdot \nabla_\theta V^{G_\theta}(Y_{1:1})] \nonumber\\
& = \sum_{y_1 \in \mathcal{Y}} \nabla_\theta G_\theta(y_1|s_0) \cdot Q^{G_\theta}(s_0,y_1) + \sum_{y_1 \in \mathcal{Y}} G_\theta(y_1|s_0)\nabla_\theta[\sum_{y_2 \in \mathcal{Y}} G_\theta(y_2|Y_{1:1}) Q^{G_\theta}(Y_{1:1},y_2)] \nonumber\\
& = \sum_{y_1 \in \mathcal{Y}} \nabla_\theta G_\theta(y_1|s_0) \cdot Q^{G_\theta}(s_0,y_1) + \sum_{y_1 \in \mathcal{Y}} G_\theta(y_1|s_0) \sum_{y_2 \in \mathcal{Y}}[\nabla_\theta G_\theta(y_2|Y_{1:1})\cdot Q^{G_\theta}(Y_{1:1},y_2)\nonumber\\
& ~~~~~~~~~~~~~~~~~~~~~~~~~~~~~~~~~~~~~~~~~~~~~~~~~~~~~~
~~~~~~~~~~~~~~~~~~~~+G_\theta(y_2|Y_{1:1}) \nabla_\theta Q^{G_\theta}(Y_{1:1},y_2)] \nonumber\\
& = \sum_{y_1 \in \mathcal{Y}} \nabla_\theta G_\theta(y_1|s_0) \cdot Q^{G_\theta}(s_0,y_1) + \sum_{Y_{1:1}}P(Y_{1:1}|s_0;G_\theta) \sum_{y_2 \in \mathcal{Y}} \nabla_\theta G_\theta(y_2|Y_{1:1})\cdot Q^{G_\theta}(Y_{1:1},y_2) \nonumber\\
& ~~~~~~~~~~~~~~~~~~~~~~~~~~~~~~~~~~~~~~~~~~~~~~~~~~~~~~~~~~~~~~~~~~~~~~~~~+ \sum_{Y_{1:2}}P(Y_{1:2}|s_0;G_\theta)\nabla_\theta V^{G_\theta}(Y_{1:2}) \nonumber\\
& = \sum_{t=1}^{T} \sum_{Y_{1:t-1}}P(Y_{1:t-1}|s_0;G_\theta) \sum_{y_t \in \mathcal{Y}} \nabla_\theta G_\theta(y_t|Y_{1:t-1})\cdot Q^{G_\theta}(Y_{1:t-1},y_t)  \nonumber\\
& = \sum_{t=1}^{T} \mathbb{E}_{Y_{1:t-1} \sim G_\theta}[\sum_{y_t \in \mathcal{Y}} \nabla_\theta G_\theta(y_t|Y_{1:t-1})\cdot Q^{G_\theta}(Y_{1:t-1},y_t)],
\end{align}
}
which is the result in Eq.~(6) of the paper.

\subsection{Model Implementations}
In this section, we present a full version of the discussed generative model and discriminative model in our paper submission.

\subsubsection{The Generative Model for Sequences}
We use recurrent neural networks (RNNs) \cite{hochreiter1997long} as the generative model. An RNN maps the input embedding representations $\bs{x}_1,\ldots,\bs{x}_T$ of the sequence $x_1,\ldots,x_T$ into a sequence of hidden states $\bs{h}_1,\ldots,\bs{h}_T$ by using the update function $g$ recursively.
\begin{equation}
\bs{h}_t = g(\bs{h}_{t-1},\bs{x}_t)
\end{equation}

Moreover, a softmax output layer $z$ maps the hidden states into the output token distribution
\begin{equation}
p(y_t|x_1,\ldots,x_t) = z(\bs{h}_t) = \softmax(\bs{c}+\bs{Vh}_t),
\end{equation}
where the parameters are a bias vector $\bs{c}$ and a weight matrix $\bs{V}$.

The vanishing and exploding gradient problem in backpropagation through time (BPTT) issues a challenge of learning long-term dependencies to recurrent neural network \cite{goodfellow2016deep}. To address such problems, gated RNNs have been designed based on the basic idea of creating paths through time that have derivatives that neither vanish nor explode.
Among various gated RNNs, we choose the Long Short-Term Memory (LSTM) \cite{hochreiter1997long} to be our generative networks with the update equations:
\begin{equation}
\begin{split}
\bs{f}_t &= \sigma(\bs{W}_f\cdot[\bs{h}_{t-1},\bs{x}_t]+\bs{b}_f), \\
\bs{i}_t &= \sigma(\bs{W}_i\cdot[\bs{h}_{t-1},\bs{x}_t]+\bs{b}_i), \\
\bs{o}_t &= \sigma(\bs{W}_o\cdot[\bs{h}_{t-1},\bs{x}_t]+\bs{b}_o),\\
\bs{s}_t &= \bs{f}_t \odot \bs{s}_{t-1}+\bs{i}_t \odot \tanh(\bs{W}_s\cdot[\bs{h}_{t-1},\bs{x}_t]+\bs{b}_s), \\
\bs{h}_t &= \bs{o}_t \odot \tanh(\bs{s}_t),
\end{split}
\end{equation}
where $[\bs{h},\bs{x}]$ is the vector concatenation and $\odot$ is the elementwise product.

For simplicity, we use the standard LSTM as the generator, while it is worth noticing that most of the RNN variants, such as the gated recurrent unit (GRU) \cite{cho2014learning} and soft attention mechanism \cite{bahdanau2014neural}, can be used as a generator in SeqGAN.


The standard way of training an RNN $G_\theta$ is the maximum likelihood estimation (MLE), which involves minimizing the negative log-likelihood $-\sum_{t=1}^{T} \log G_\theta(y_t=x_t|\left\{x_1,\ldots,x_{t-1}\right\})$ for a generated sequence $(y_1,\ldots,y_T)$ given input $(x_1,\ldots,x_T)$.
However, when applying MLE to generative models, there is a discrepancy between training and generating \cite{bengio2015scheduled,huszar2015not}, which motivates our work.

\subsubsection{The Discriminative Model for Sequences}

Deep discriminative models such as deep neural network (DNN) \cite{vesely2013sequence}, convolutional neural network (CNN) \cite{kim2014convolutional} and recurrent convolutional neural network (RCNN) \cite{lai2015recurrent} have shown a high performance in complicated sequence classification tasks. In this paper, we choose the CNN as our discriminator as CNN has recently been shown of great effectiveness in text (token sequence) classification \cite{zhang2015text}.

As far as we know, except for some specific tasks, most discriminative models can only perform classification well for a whole sequence rather than the unfinished one. In case of some specific tasks, one may design a classifier to provide intermediate reward signal to enhance the performance of our framework. But to make it more general, we focus on the situation where discriminator can only provide final reward, i.e., the probability that a finished sequence was real.


We first represent an input sequence $x_1,\ldots,x_T$ as:
\begin{equation}
\mathcal{E}_{1:T} = \bs{x}_1 \concat \bs{x}_2 \concat \ldots \concat \bs{x}_T,
\end{equation}
where $\bs{x}_t\in \mathbb{R}^k$ is the $k$-dimensional token embedding and $\concat$ is the vertical concatenation operator to build the matrix $\mathcal{E}_{1:T} \in \mathbb{R}^{T\times k}$. Then a kernel $\bs{w} \in \mathbb{R}^{l \times k}$ applies a convolutional operation to a window size of $\mathnormal{l}$ words to produce a new feature map:
\begin{equation}
c_i = \rho(\bs{w}\otimes \mathcal{E}_{i:i+l-1}+b),
\end{equation}
where $\otimes$ operator is the summation of elementwise production, $\mathnormal{b}$ is a bias term and $\rho$ is a non-linear function.
We can use various numbers of kernels with different window sizes to extract different features. Specifically, a kernel $\bs{w}$ with window size $l$ applied to the concatenated embeddings of input sequence will produce a feature map
\begin{equation}
\bs{c} = [c_1,\ldots,c_{T-l+1}].
\end{equation}

Finally we apply a max-over-time pooling operation over the feature map $\tilde{c} = \max\left\{\bs{c}\right\}$ and pass all pooled features from different kernels to a fully connected softmax layer to get the probability that a given sequence is real.

We perform an empirical experiment to choose the kernel window sizes and numbers as shown in Table \ref{tab:hyper}. For different tasks, one should design specific structures for the discriminator.
\begin{table}[h]
	\centering
	\vspace{-5pt}
	\caption{Convolutional layer structures.}\label{tab:hyper}
	\vspace{2pt}
	\small
	\begin{tabular}{c|c}
		Sequence length & (window size, kernel numbers) \\
		\hline
		\multirow{3}*{20} & (1, 100),(2, 200),(3, 200),(4, 200),(5, 200)\\
		&(6, 100),(7, 100),(8, 100),(9, 100),(10, 100)\\
		&(15, 160),(20, 160) \\
		\hline
		\multirow{3}*{32} & (1, 100),(2, 200),(3, 200),(4, 200),(5, 200)\\
		   & (6, 100),(7, 100),(8, 100),(9, 100),(10, 100)\\
		   & (16, 160),(24, 160),(32, 160)\\

	\end{tabular}
\end{table}

To enhance the performance, we also add the highway architecture \cite{srivastava2015highway} before the final fully connected layer:
\begin{align}\label{eq:hw}
\begin{split}
\bs\tau &= \sigma(\bs{W}_T \cdot \tilde{\bs{c}}+ \bs{b}_T),\\
\tilde{\bs{C}} &= \bs\tau \cdot H(\tilde{\bs{c}},\bs{W}_H) + (\bs{1} - \bs\tau) \cdot \tilde{\bs{c}},
\end{split}
\end{align}
where $\bs{W}_T$, $\bs{b}_T$ and $\bs{W}_H$ are highway layer weights, $H$ denotes an affine transform followed by a non-linear activation function such as a rectified linear unit (ReLU) and $\bs\tau$ is the ``transform gate" with the same dimensionality as $H(\tilde{\bs{c}},\bs{W}_H)$ and $\tilde{\bs{c}}$.
Finally, we apply a sigmoid transformation to get the probability that a given sequence is real:
\begin{align}
\hat{y} = \sigma(\bs{W}_o \cdot \tilde{\bs{C}} + b_o)
\end{align}
where $\bs{W}_o$ and $b_o$ is the output layer weight and bias.

When optimizing discriminative models, supervised training is applied to minimize the cross entropy, which is widely used as the objective function for classification and prediction tasks:
\begin{equation}
\mathcal{L}(y,\hat{y}) = -y \log\hat{y} - (1-y) \log(1-\hat{y}), \label{eq:classification-loss}
\end{equation}
where $y$ is the ground truth label of the input sequence and $\hat{y}$ is the predicted probability from the discriminative models.


\subsection{More Ablation Study}
\begin{figure}[h]
\centering
\includegraphics[width=0.6\columnwidth]{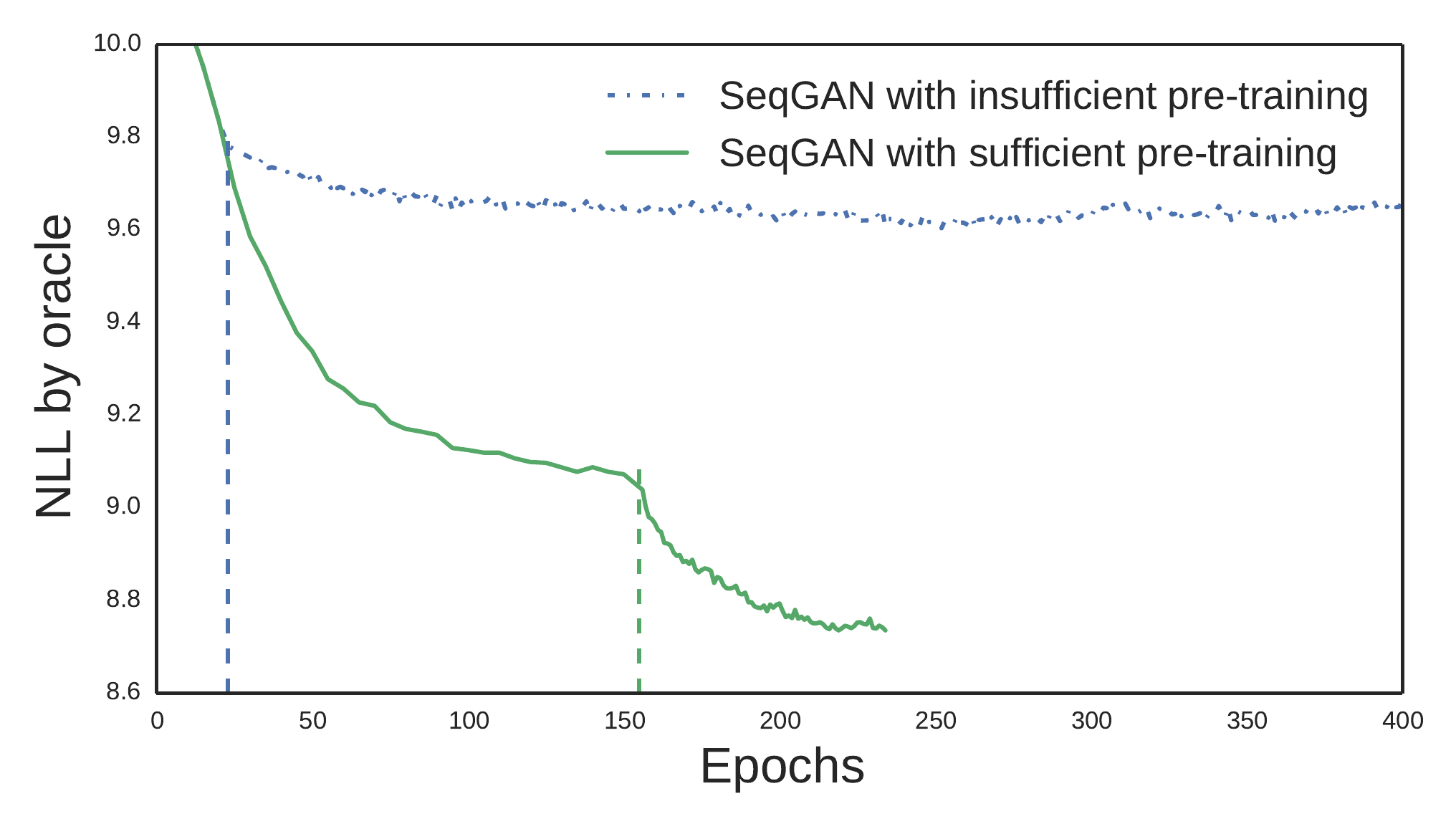}
\caption{Negative log-likelihood performance with different pre-training epochs before the adversarial training. The vertical dashed lines represent the start of adversarial training.}
\label{fig:learning-curve}
\end{figure}

In the \textsc{Discussion} subsection of \textsc{Synthetic Data Experiments} section of our paper, we discussed the ablation study of three hyperparameters of SeqGAN, i.e., g-steps, d-steps and $k$ epoch number. Here we provide another ablation study which is instructive for the better training of SeqGAN.

As described in our paper, we start the adversarial training process after the convergence of MLE supervised pre-training. Here we further conduct experiments to investigate the performance of SeqGAN when the supervised pre-training is insufficient.

As shown in Figure \ref{fig:learning-curve}, if we pre-train the generative model with conventional MLE methods for only 20 epochs, which is far from convergence, then the adversarial training process improves the generator quite slowly and unstably. The reason is that in SeqGAN, the discriminative model provides reward guidance when training the generator and if the generator acts almost randomly, the discriminator will identify the generated sequence to be unreal with high confidence and almost every action the generator takes receives a low (unified) reward, which does not guide the generator towards a good improvement direction, resulting in an ineffective training procedure. This indicates that in order to apply adversarial training strategies to sequence generative models, a sufficient pre-training is necessary.

\end{appendices}
\end{document}